\documentclass[11pt]{article}

\usepackage{acl}

\usepackage{times}
\usepackage{latexsym}
\usepackage[T1]{fontenc}
\usepackage[utf8]{inputenc}
\usepackage{microtype}
\usepackage{inconsolata}
\usepackage{graphicx}
\usepackage{amsmath}
\usepackage{amssymb}
\usepackage{booktabs}
\usepackage{multirow}
\usepackage{dblfloatfix}

\title{LLM Pedagogical Behavior in AI Tutoring Interactions}

\author{
  Suhyeon Lee \\
  KAIST \\
  \texttt{suhyeonlee@kaist.ac.kr} \\\And
  Juneha Baek \\
  KAIST \\
  \texttt{juneha.baek@kaist.ac.kr} \\\AND
  Jaehyeong Park \\
  KAIST \\
  \texttt{hyeong@kaist.ac.kr} \\\And
  Donghyuk Shin\thanks{\ Corresponding author.} \\
  KAIST \\
  \texttt{dhs@kaist.ac.kr} \\
}

\begin{document}
\maketitle

\begin{abstract}
Students increasingly use LLMs as tutors for coursework and problem solving.
Little is known about the level of assistance LLMs provide when students use them as tutors in authentic learning interactions.
This matters because tutoring responses can differ substantially in how directly they help students complete a task.
We operationalize this dimension as scaffolding level and develop a five-level scale, validated against human annotations, that characterizes responses according to the degree of direct assistance they provide.
We apply the scale to 14,637 LLM responses from 203 students in a university AI course.
Responses are overwhelmingly concentrated at high levels of assistance, with more than 95\% classified as either Explaining or Solving.
Scaffolding level is systematically associated with students' subsequent conversational behavior, but provides little additional predictive information about performance on three subsequent exams beyond prior achievement and dialogue behavior.
These findings provide an empirical baseline for LLM assistance in tutoring interactions and a measurement framework for evaluating how alternative tutoring designs change that assistance.
\end{abstract}

\section{Introduction}

Large language models (LLMs) are increasingly used by students as on-demand tutors for programming assignments and other learning tasks \citep{ghimire2024coding, ma2024enhancing}.
Students can ask for clarification, explanations, hints, or complete solutions, and LLM responses can vary substantially in how directly they help students complete a task.
Yet little is known about the level of assistance LLMs provide when students use them as tutors in authentic learning interactions.
Understanding this behavior is important because different levels of assistance imply different roles for the learner in working through the task.

This distinction is central to theories of instructional support.
The Assistance Dilemma \citep{koedinger2007exploring} emphasizes the tension between providing too little help, which may impede progress, and too much help, which may bypass productive cognitive effort.
Cognitive Apprenticeship similarly describes learning as a gradual transfer of responsibility from instructor to learner \citep{collins1989cognitive}.
From this perspective, tutoring responses differ not only in what information they contain, but also in the degree of direct assistance they provide.
We refer to this dimension as \textit{scaffolding level}.

Existing research provides important evidence about LLM-assisted learning but has primarily focused on other aspects of the interaction.
Observational studies examine what students ask, how they use LLMs, and how usage relates to course outcomes \citep{ghimire2024coding, ma2024enhancing, brender2024whos, mcnichols2026studychat}.
A separate line of work explicitly designs LLM-based tutors to promote reasoning or structure how assistance is delivered \citep{liu2024socraticlm, learnlm2024, peng2025kele}.
What remains less well characterized is the assistance that students actually receive across authentic LLM tutoring interactions.
Establishing this behavioral baseline is useful both for understanding current LLM tutoring behavior and for evaluating whether alternative tutoring designs meaningfully change it.

We address this gap by developing a five-level scaffolding scale that characterizes LLM responses according to the degree of direct assistance they provide.
We validate the scale against human annotations and apply it to 14,637 LLM responses from 203 students in the StudyChat dataset \citep{mcnichols2026studychat}, collected in an introductory university AI course where the model received no pedagogical instruction beyond ``You are a helpful assistant.''
Because the dataset links LLM responses to subsequent student turns and course assessments, we examine not only the distribution of scaffolding levels but also whether this variation is reflected in subsequent interaction and later performance.

We ask:

\begin{itemize}
  \item \textbf{RQ1}: What scaffolding levels does the LLM exhibit, and how do they vary across student requests and assignments?
  \item \textbf{RQ2}: Is the LLM's scaffolding level associated with what the student does next?
  \item \textbf{RQ3}: Does scaffolding provide additional predictive information about subsequent assessment performance?
\end{itemize}

We find that the assistant operates predominantly at high levels of assistance: more than 95\% of responses are classified as either Explaining or Solving.
Scaffolding level is systematically associated with students' subsequent conversational behavior, including within broad categories of student requests.
By contrast, it provides little additional predictive information about performance on three subsequent exams beyond prior achievement and observed dialogue behavior.
Together, these findings provide an empirical baseline for LLM assistance in authentic tutoring interactions and a measurement framework for evaluating how alternative tutoring designs alter that behavior.

\section{Related Work}

\subsection{Student Use of LLMs in Education}

Research on LLM-assisted learning has largely focused on student behavior.
\citet{ghimire2024coding} find that students in programming courses more often ask questions than directly request solutions, while \citet{ma2024enhancing} report similar patterns in a Python course.
\citet{brender2024whos} examine how ChatGPT use relates to learning activities.
\citet{mcnichols2026studychat} classify student utterances in StudyChat into dialogue acts and relate those interaction patterns to course outcomes.

This work provides an increasingly detailed account of the student side of LLM-assisted learning.
Our focus is complementary: rather than characterizing what students ask, we characterize the amount of assistance contained in the LLM response.

\subsection{Tutoring Dialogue and Scaffolding}

Research on tutoring dialogue has long examined patterns of communicative moves between tutors and learners \citep{graesser1995collaborative, graesser2004autotutor, boyer2011affect, vail2014identifying}.
More recent work uses LLM-based annotation to scale dialogue analysis in educational settings \citep{wang2023can, yin2025scaling}.
Dialogue-act schemes generally characterize the communicative function of an utterance within an interaction.

Scaffolding captures a different dimension: how much assistance a response provides.
Cognitive Apprenticeship describes instruction as progressively transferring responsibility from instructor to learner \citep{collins1989cognitive, dennen2004cognitive}, while the Assistance Dilemma emphasizes balancing guidance with opportunities for productive effort \citep{koedinger2007exploring}.
We operationalize this dimension as a continuum ranging from responses that provide little substantive assistance to responses that directly solve the student's task.

\subsection{Pedagogically Designed LLM Tutors}

A growing line of work explicitly designs LLMs to behave as pedagogical tutors rather than general-purpose assistants.
SocraticLM replaces a question-answering paradigm with a Socratic tutoring paradigm designed to engage students in the reasoning process \citep{liu2024socraticlm}.
LearnLM formulates pedagogical behavior as instruction following and incorporates pedagogical examples into model post-training, allowing desired teaching behaviors to be specified explicitly \citep{learnlm2024}.
KELE combines a structured Socratic teaching rule system with a multi-agent architecture that separates pedagogical planning from response generation \citep{peng2025kele}.

These approaches demonstrate that LLM tutoring behavior can be deliberately shaped through pedagogical design.
Our question is complementary and logically prior: what assistance emerges when students use a general-purpose LLM without such a tutoring strategy?
Characterizing this baseline provides a reference point for evaluating whether pedagogically designed systems change the assistance students actually receive.

\section{Data}

We use the StudyChat dataset \citep{mcnichols2026studychat}, collected from an introductory AI course at a U.S.\ university across Fall 2024 and Spring 2025.
Students interacted with a ChatGPT-like interface backed by GPT-4o-mini.
The system prompt was ``You are a helpful assistant.''
No additional instruction specified how the model should teach or how much assistance it should provide.

The dataset contains 16,851 turns from 203 students across 2,214 conversations and seven programming assignments organized into three course modules (Table~\ref{tab:modules}).

\begin{table}[t]
  \centering
  \small
  \begin{tabular}{llll}
    \toprule
    \textbf{Module} & \textbf{Topic} & \textbf{Assignments} & \textbf{Exam} \\
    \midrule
    1 & Search algorithms   & a1, a2       & e1 \\
    2 & Machine learning    & a3, a4, a5   & e2 \\
    3 & Applied AI          & a6, a7       & e3 \\
    \bottomrule
  \end{tabular}
  \caption{Course structure. Assignments allow LLM use; exams do not.}
  \label{tab:modules}
\end{table}

Student utterances include dialogue-act (DA) labels provided with the dataset.
These labels were generated through LLM annotation and validated against human annotations, with $\kappa = .74$ at the broad level.
We use the broad categories throughout; the full taxonomy is reported in Appendix~\ref{sec:additional}.
Assessment data are available for 181 students.
Students could use the LLM while completing assignments, whereas the three exams were completed without LLM access.

\section{Methods}

\subsection{Paired Turn Construction}

We construct interaction units $(s_t, d_t, r_t, d_{t+1})$, where $s_t$ is the student prompt at turn $t$, $d_t$ is its broad dialogue-act label, $r_t$ is the corresponding LLM response, and $d_{t+1}$ is the dialogue act of the student's next utterance.
We retain the two preceding turns as context for response classification.
Restricting the sample to LLM responses followed by a student turn yields 14,637 paired turns.
Of these, 12,692 come from students with assessment data.

\subsection{Measuring Scaffolding Level}

We classify each LLM response on a five-level ordinal scale according to the amount of assistance it provides: L0 (Minimal) provides no substantive task-relevant assistance; L1 (Prompting) asks the student to reason or try without directing them toward a solution; L2 (Hinting) identifies a relevant concept, function, or error location without providing a concrete solution; L3 (Explaining) explains an approach that the student must still apply; and L4 (Solving) provides a complete, task-specific solution that can be used directly.

Higher levels provide more directly usable assistance and leave less of the task for the student to complete independently.
Let $L_t \in \{0,1,2,3,4\}$ denote the scaffolding level assigned to response $r_t$.

Responses are classified using GPT-4o-mini at temperature 0.
The classifier receives the category definitions, the target response, and the preceding context.
For RQ1, we summarize scaffolding levels overall and across student dialogue acts and assignments.
We evaluate assignment-level differences using a Kruskal--Wallis test.

\paragraph{Human validation.}
We validate the scale on a stratified sample of 300 responses.
Two authors independently annotate all responses while blind to each other's labels and to the classifier outputs.
Disagreements are resolved through discussion.
Human-human agreement is high ($\kappa = .901$; 95.0\% raw agreement), including at the L3/L4 boundary ($\kappa = .894$).
Against the adjudicated human labels, the classifier achieves 75.3\% accuracy and a weighted $F_1$ of .751 on the held-out test split ($n = 150$).
Most errors occur between L3 and L4.
Across the 300 responses, 57 human-labeled L4 responses are classified as L3, against 9 in the reverse direction.
The classifier therefore shows greater under-detection than over-detection of Solving responses.
The full codebook, confusion matrix, and a sensitivity analysis collapsing the L3/L4 boundary are reported in Appendix~\ref{sec:classification}.

\subsection{Scaffolding and Student Follow-up}

For each scaffolding level $l$, we compute the conditional distribution of the next student dialogue act:
\begin{equation}
  P(d_{t+1} {=} j \mid L_t {=} l)
  =
  \frac{|\{t : L_t {=} l \wedge d_{t+1} {=} j\}|}
       {|\{t : L_t {=} l\}|}.
\end{equation}
We test the association between $L_t$ and $d_{t+1}$ using a chi-squared test of independence.
Because different student requests may elicit both different LLM responses and different subsequent behaviors, we repeat the analysis within each broad current dialogue-act category $d_t$.
This analysis tests whether the association persists beyond broad differences in request type.

\subsection{Scaffolding and Assessment Performance}

We examine whether scaffolding provides additional predictive information about subsequent exam performance.
For each exam $a$ (e1--e3), we estimate:
\begin{equation}
  y_{i,a}
  =
  \beta_0
  + \beta_1 \bar{y}_{i,{<}a}
  + \textstyle\sum_k \gamma_k \, \text{DA}_{i,k,{<}a}
  + \delta \, \bar{L}_{i,{<}a}
  + \epsilon_{i,a}.
\end{equation}
where $y_{i,a}$ is student $i$'s normalized score on exam $a$, $\bar{y}_{i,{<}a}$ is the student's mean prior assessment performance, $\text{DA}_{i,k,{<}a}$ is the cumulative count of student dialogue act $k$ before exam $a$, and $\bar{L}_{i,{<}a}$ is the mean scaffolding level received before exam $a$.
We compare a baseline model containing prior performance and dialogue-act counts with a full model that additionally includes mean scaffolding level.
The comparison measures the incremental predictive information provided by scaffolding beyond prior achievement and observed dialogue behavior.

\section{Results}

\subsection{RQ1: What Assistance Does the LLM Provide?}

The response distribution is strongly concentrated at high levels of assistance (Table~\ref{tab:level_distribution}).

\begin{table}[t]
  \centering
  \small
  \begin{tabular}{llrr}
    \toprule
    \textbf{Level} & \textbf{Description} & \textbf{$n$} & \textbf{\%} \\
    \midrule
    L4 & Solving      & 8{,}085 & 55.2 \\
    L3 & Explaining   & 5{,}899 & 40.3 \\
    L2 & Hinting      &      43 &  0.3 \\
    L1 & Prompting    &     288 &  2.0 \\
    L0 & Minimal      &     322 &  2.2 \\
    \bottomrule
  \end{tabular}
  \caption{Distribution of scaffolding levels ($N = 14{,}637$).}
  \label{tab:level_distribution}
\end{table}

More than half of responses provide a complete, task-specific solution, and another 40.3\% provide detailed explanations.
Prompting, Hinting, and Minimal responses together account for fewer than 5\% of responses.
The mean scaffolding level is $\bar{L} = 3.43$ (SD = 0.84).
Thus, the dominant response pattern is Explaining or Solving rather than Prompting or Hinting.

Assistance remains high across the major instructional request types.
Writing requests receive the highest mean scaffolding level ($\bar{L} = 3.72$), followed by editing requests (3.57), provide-context turns (3.53), conceptual questions (3.38), verification requests (3.34), and contextual questions (3.27).
Off-topic turns differ sharply ($\bar{L} = 0.87$), reflecting the absence of a substantive instructional task.

\begin{figure}[t]
  \centering
  \includegraphics[width=\columnwidth]{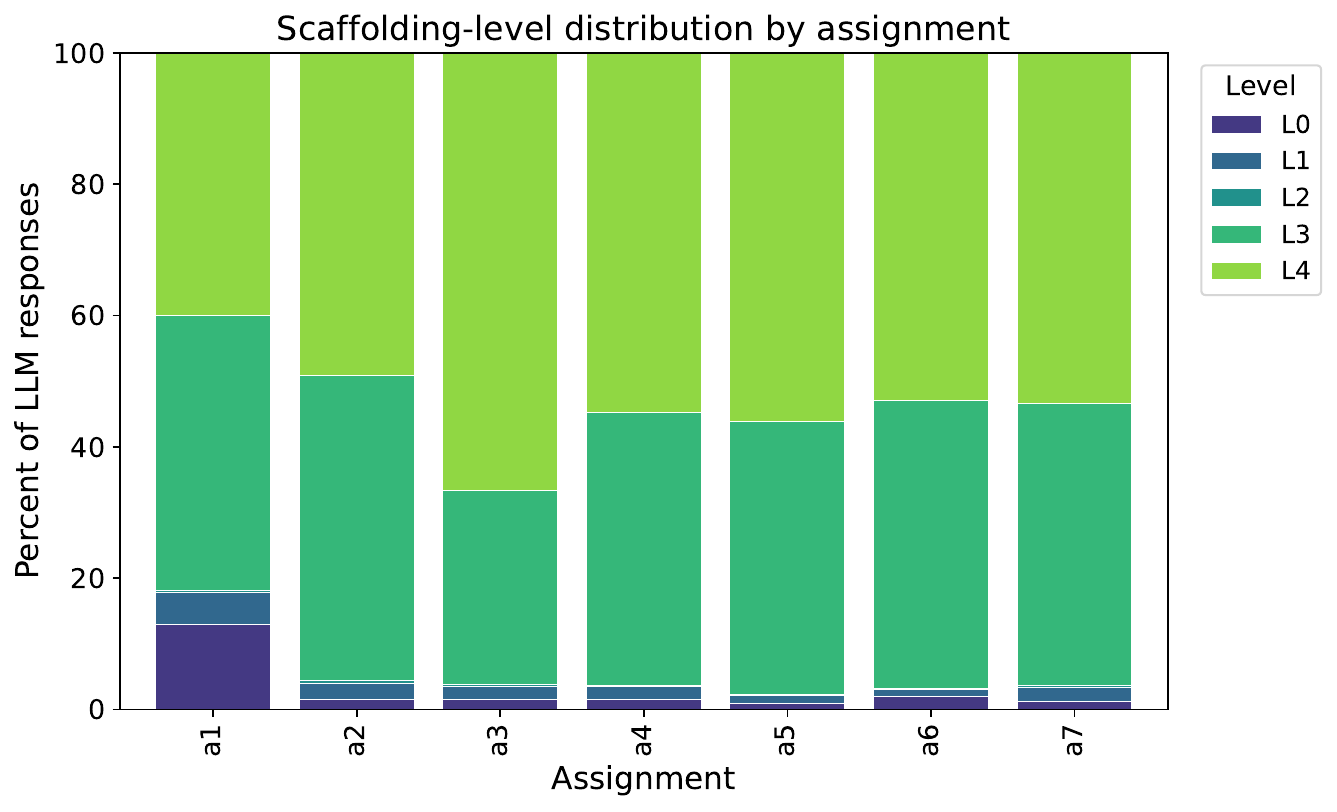}
  \caption{Distribution of scaffolding levels by assignment.}
  \label{fig:assignment_level}
\end{figure}

Scaffolding also varies across assignments (Kruskal--Wallis $H = 336.9$, $p < .001$; Figure~\ref{fig:assignment_level}).
For example, the pandas assignment a3 has a mean scaffolding level of 3.58, compared with 2.91 for the search-algorithm assignment a1.
Despite this variation, high-assistance responses remain dominant across assignments.

\subsection{RQ2: Is Scaffolding Associated with What Students Do Next?}

The distribution of students' next dialogue acts differs across scaffolding levels ($\chi^2 = 1{,}092.6$, $df = 28$, $p < .001$, $V = .137$; Table~\ref{tab:transition}).

\begin{table}[t]
  \centering
  \small
  \begin{tabular}{lrrrrr}
    \toprule
    & \multicolumn{4}{c}{\textbf{Next student DA (\%)}} & \\
    \cmidrule(lr){2-5}
    \textbf{Level} & \textbf{ConQ} & \textbf{WrReq} & \textbf{PrCtx} & \textbf{CtxQ} & \textbf{$n$} \\
    \midrule
    L0 & 13.7 & 25.8 & 14.0 & 23.6 &    322 \\
    L1 &  9.7 & 24.3 & 34.0 & 15.3 &    288 \\
    L2 & 39.5 & 11.6 & 23.3 & 18.6 &     43 \\
    L3 & 40.5 & 19.2 & 13.6 & 19.8 & 5{,}899 \\
    L4 & 26.0 & 29.7 & 20.6 & 16.3 & 8{,}085 \\
    \bottomrule
  \end{tabular}
  \caption{Distribution of the next student dialogue act (\%) by LLM scaffolding level. ConQ: conceptual question, WrReq: writing request, PrCtx: provide context, CtxQ: contextual question. Minor categories are omitted; definitions appear in Appendix~\ref{sec:additional}.}
  \label{tab:transition}
\end{table}

The largest scaffolding categories show distinct follow-up patterns.
After Solving responses, students most often make another writing request (29.7\%).
After Explaining responses, conceptual questions are most common (40.5\%).
After Prompting responses, students most often provide additional context (34.0\%).

The association remains statistically significant within all four broad current student dialogue-act categories examined (all $p < .01$).
The pattern therefore persists beyond broad differences in the type of request preceding the LLM response.
The association also remains when Explaining and Solving are collapsed into a single high-assistance category, although the effect size decreases from $V = .137$ to $V = .116$.
Detailed within-dialogue-act analyses and additional follow-up measures are reported in Appendix~\ref{sec:additional}.

\subsection{RQ3: Does Scaffolding Add Predictive Information About Exam Performance?}

Assignment grades exhibit strong ceiling effects, so we focus on the three exams, which show greater score variation and were completed without LLM access.

\begin{table}[t]
  \centering
  \small
  \begin{tabular}{lrrrrrl}
    \toprule
    \textbf{Exam} & $n$ & $R^2_{\text{base}}$ & $R^2_{\text{full}}$ & $\Delta R^2$ & $\hat{\delta}$ & $p_F$ \\
    \midrule
    e1 & 147 & .123 & .143 & +.021 & +.020  & .072 \\
    e2 & 173 & .303 & .314 & +.011 & +.021  & .104 \\
    e3 & 173 & .240 & .250 & +.010 & $-.019$ & .149 \\
    \bottomrule
  \end{tabular}
  \caption{Incremental prediction of exam scores from mean scaffolding level. Baseline models include prior performance and student dialogue-act counts; full models additionally include mean scaffolding level.}
  \label{tab:regression}
\end{table}

The baseline models explain approximately 12--30\% of exam-score variance.
Adding mean scaffolding increases explained variance by only 1--2 percentage points.
None of the three model comparisons reaches conventional statistical significance (Table~\ref{tab:regression}), and the estimated scaffolding coefficient changes direction for e3.
Scaffolding therefore provides no consistent additional predictive information about subsequent exam performance beyond prior achievement and observed dialogue behavior.
Alternative specifications using the frequency of Solving responses or low-assistance responses yield the same overall pattern (Appendix~\ref{sec:alt_regression}).

\section{Discussion}

\subsection{Default LLM Assistance Is Highly Direct}

A general-purpose assistant used without explicit pedagogical prompting overwhelmingly provides substantial assistance.
More than 95\% of responses are classified as either Explaining or Solving, while Prompting and Hinting are rare.
Although assistance varies across student requests and assignments, high-assistance responses remain dominant.

This distribution provides a behavioral baseline for general-purpose LLM tutoring.
From the perspective of the Assistance Dilemma, assistance level matters because it determines how much of the task remains with the learner.
In the setting studied here, general helpfulness manifests primarily as explanation and direct solution provision rather than prompting or hinting.
The finding establishes a baseline for comparison rather than identifying an optimal assistance level.

\subsection{Scaffolding Is Reflected in Subsequent Interaction}

Scaffolding level is systematically associated with how students continue the conversation.
Explaining responses are followed relatively often by conceptual questions, whereas Solving responses are followed more often by additional writing requests.
These patterns persist within broad categories of initial student requests and after collapsing the L3/L4 distinction.

Scaffolding therefore captures variation in LLM responses that is reflected in subsequent student interaction.
At the same time, the observational design does not identify whether scaffolding itself causes these differences.
Prompt content, task difficulty, conversation history, and student characteristics may influence both the response received and the behavior that follows.

\subsection{Interactional Relevance Does Not Imply Later Performance Differences}

The assessment results distinguish immediate interactional patterns from later academic outcomes.
Although scaffolding is associated with students' next conversational moves, it adds little predictive information about subsequent exam performance once prior achievement and observed dialogue behavior are included.

This pattern does not imply that assistance level has no effect on learning.
The assessment sample is modest, and responses are heavily concentrated at L3 and L4, leaving limited variation across the full scaffolding continuum.
The present data therefore provide clear evidence about the assistance the LLM provides and how that variation corresponds to subsequent interaction, but not about the causal learning effects of different scaffolding strategies.

\subsection{Implications for Evaluating LLM Tutors}

The findings suggest that evaluations of LLM tutoring systems should distinguish two questions.
The first is behavioral: does a pedagogical intervention change the assistance students receive?
The second is educational: does that change improve learning?

The scaffolding measure provides a way to address the first question directly.
A tutoring prompt, policy, or model intended to encourage greater student reasoning should produce a measurable shift in scaffolding relative to the general-purpose baseline.
Whether such a shift improves learning is a separate empirical question.

\section*{Limitations}

The study has four main limitations.

First, the data come from one introductory AI course and one model configuration.
The observed distribution therefore characterizes this setting rather than general-purpose LLMs universally.
Here, ``default'' refers to use without explicit pedagogical prompting.

Second, automated scaffolding classification is imperfect despite high human agreement on the underlying scale.
Most errors occur at the L3/L4 boundary, and low-assistance levels are rare in both the full dataset and the human validation sample.
Fine-grained estimates for L0--L2 should therefore be interpreted cautiously.

Third, the follow-up analyses are observational.
Conditioning on broad student dialogue-act categories does not account for all differences in prompt content, task difficulty, student characteristics, or conversation history.
Students also contribute multiple turns, whereas the turn-level chi-squared analyses do not explicitly model within-student dependence.

Finally, the assessment analysis evaluates incremental prediction rather than the causal effect of scaffolding on learning.
The assessment sample is modest, and the strong concentration of responses at L3 and L4 limits the variation available for detecting differences in subsequent performance.

\section*{Ethics Statement}

This study uses the publicly released StudyChat dataset \citep{mcnichols2026studychat}, collected under IRB approval with informed consent at the University of Massachusetts Amherst, with personally identifiable information removed by the original authors before release.
Our use is consistent with its stated research purpose.

The 300-item validation set was annotated by two of the authors, each working independently and blind to the other's labels and to the classifier outputs.
No external annotators were recruited.

Classification used the OpenAI API (GPT-4o-mini).
AI writing tools were used for drafting and editing, and all content was reviewed and verified by the authors.

\section*{Acknowledgments}

This work was supported by the Institute of Information \& Communications Technology Planning \& Evaluation (IITP)-Global Data-X Leader HRD program grant funded by the Korea government (MSIT) (IITP-RS-2024-00440626).

\bibliography{custom}

\appendix

\section{Scaffolding Measurement Details}
\label{sec:classification}

\subsection{Scaffolding Codebook}

We classify each LLM response according to the amount of assistance it provides and the amount of task completion left to the student.
Table~\ref{tab:codebook} presents the full codebook.

\begin{table*}[!tb]
  \centering
  \small
  \begin{tabular}{p{0.06\textwidth}p{0.10\textwidth}p{0.40\textwidth}p{0.34\textwidth}}
    \toprule
    \textbf{Level} & \textbf{Label} & \textbf{Definition} & \textbf{Example} \\
    \midrule
    L0 & Minimal & Provides no substantive task-relevant assistance. Includes refusals, out-of-scope responses, or responses that redirect the question without useful guidance. & ``I can't complete your assignment for you.'' \\
    \addlinespace
    L1 & Prompting & Asks the student to reason or try first without directing them toward a solution. & ``What do you think happens when the learning rate is too high?'' \\
    \addlinespace
    L2 & Hinting & Identifies a relevant concept, function, or error location without providing a concrete solution. & ``The issue is likely in your loss function. Consider how cross-entropy handles multi-class labels.'' \\
    \addlinespace
    L3 & Explaining & Explains a concept, algorithm, or general approach in enough detail for the student to apply it, without providing a directly usable task-specific solution. & An explanation of how cross-validation works and why $k$-fold validation may be appropriate. \\
    \addlinespace
    L4 & Solving & Provides a complete, task-specific solution that the student can use directly. & A full Python function implementing uniform-cost search using the student's variables and assignment context. \\
    \bottomrule
  \end{tabular}
  \caption{Scaffolding-level codebook.}
  \label{tab:codebook}
\end{table*}

The main distinction between L3 and L4 is direct usability.
L3 provides an approach that the student must still adapt or implement, whereas L4 applies the approach to the student's specific task in a form that can be used directly.

\subsection{Human Validation}

We draw a stratified sample of 300 responses across assignments, with oversampling designed to increase representation of the rare low-assistance levels.
Two authors independently annotate all 300 responses using the codebook in Table~\ref{tab:codebook}, blind to each other's labels and to the classifier outputs.
Disagreements are resolved through discussion, and the adjudicated labels form the human reference set.

Human-human agreement is high, with Cohen's $\kappa = .901$ and 95.0\% raw agreement.
Agreement remains similarly high at the L3/L4 boundary ($\kappa = .894$).

Against the human reference labels, the GPT-4o-mini classifier achieves 75.3\% accuracy, weighted $F_1 = .751$, and Cohen's $\kappa = .527$ on the held-out test split ($n = 150$).
Across all 300 annotated responses, accuracy is 75.7\%, weighted $F_1 = .759$, and $\kappa = .552$.
Table~\ref{tab:confusion} reports the full confusion matrix.

\begin{table}[t]
  \centering
  \small
  \begin{tabular}{lrrrrrr}
    \toprule
    & \multicolumn{5}{c}{\textbf{Classifier}} & \\
    \cmidrule(lr){2-6}
    \textbf{Human} & \textbf{L0} & \textbf{L1} & \textbf{L2} & \textbf{L3} & \textbf{L4} & \textbf{$n$} \\
    \midrule
    L0 & 4 & 2 & 0 &   1 &   0 &   7 \\
    L1 & 0 & 0 & 0 &   0 &   0 &   0 \\
    L2 & 0 & 1 & 0 &   1 &   0 &   2 \\
    L3 & 0 & 0 & 0 & 107 &   9 & 116 \\
    L4 & 2 & 0 & 0 &  57 & 116 & 175 \\
    \midrule
    $n$ & 6 & 3 & 0 & 166 & 125 & 300 \\
    \bottomrule
  \end{tabular}
  \caption{Confusion matrix of human reference labels (rows) and classifier predictions (columns) for the 300 annotated responses.}
  \label{tab:confusion}
\end{table}

Because the validation sample oversamples rare response types, its marginal label frequencies should not be interpreted as estimates of the population scaffolding distribution.

\subsection{Classification Error Structure}

Most classification errors occur at the adjacent L3/L4 boundary.
Of the 73 disagreements between the human reference and classifier labels, 57 are human-labeled L4 responses classified as L3, whereas 9 are human-labeled L3 responses classified as L4.
The classifier therefore more often shifts responses from Solving to Explaining than in the reverse direction.

This error pattern affects the precise distinction between L3 and L4 but does not alter the broader concentration of responses at high assistance.
We therefore additionally examine the robustness of RQ2 to collapsing this boundary.

\subsection{Sensitivity to the L3/L4 Boundary}

We repeat the RQ2 analysis after collapsing Explaining and Solving into a single high-assistance category.
The association between scaffolding and the next student dialogue act remains significant ($\chi^2 = 592.2$, $df = 21$, $p < .001$).
Cram\'er's $V$ decreases from $.137$ in the five-level specification to $.116$ after collapsing L3 and L4.
Part of the overall association therefore reflects variation between Explaining and Solving, but the association remains when that distinction is removed.

Some cells remain sparse because L2 and several less common student dialogue acts occur infrequently.
The chi-squared $p$-values should therefore be interpreted as approximate, with greater emphasis on the observed transition patterns and effect sizes.

\subsection{Classification Prompt}

Each LLM response is classified with a single GPT-4o-mini call at temperature 0.
The model receives the scaffolding definitions, preceding conversational context, the current student request, and the response to be classified.
It returns one integer from 0 to 4 (Table~\ref{tab:prompt}).

\begin{table}[t]
  \small
  \begin{tabular}{p{0.95\columnwidth}}
    \toprule
    \textbf{System Prompt (abridged)} \\
    \midrule
    You are classifying an LLM tutoring response on a scaffolding scale (0--4). \\
    \\
    L0 (Minimal): No substantive task-relevant assistance. \\
    L1 (Prompting): Asks the student to reason or try first without providing direction toward a solution. \\
    L2 (Hinting): Identifies a relevant concept or error location without providing a concrete solution. \\
    L3 (Explaining): Explains an approach that the student must still apply. \\
    L4 (Solving): Provides a complete, task-specific solution that can be used directly. \\
    \\
    L3/L4 distinction: Determine whether the response applies the solution directly to the student's task or context. \\
    \\
    Return only one number: 0, 1, 2, 3, or 4. \\
    \midrule
    \textbf{User Prompt Template} \\
    \midrule
    {[PREV]} Student: \{previous student turn\} \\
    {[PREV]} LLM: \{previous LLM turn\} \\
    {[CLASSIFY]} Student: \{current prompt\} \\
    {[CLASSIFY]} LLM: \{current response\} \\
    \bottomrule
  \end{tabular}
  \caption{Abridged scaffolding-classification prompt.}
  \label{tab:prompt}
\end{table}

\section{Additional Results}
\label{sec:additional}

\subsection{Student Dialogue-Act Taxonomy}

The student dialogue-act labels are provided with the StudyChat dataset \citep{mcnichols2026studychat}.
The original schema contains broad categories and more specific dialogue acts.
The labels were assigned at scale through LLM annotation and validated against human annotations, with $\kappa = .74$ at the broad level.
We use the broad categories throughout our analyses.
Table~\ref{tab:da_taxonomy} lists the corresponding specific dialogue acts for reference.
The residual Misc category ($n = 35$) is excluded.

\begin{table}[t]
  \centering
  \small
  \begin{tabular}{p{0.28\columnwidth}p{0.62\columnwidth}}
    \toprule
    \textbf{Broad category} & \textbf{Specific dialogue acts} \\
    \midrule
    Writing request & Write Code, Write English, Conversion, Summarize, Other \\
    \addlinespace
    Editing request & Edit Code, Edit English, Other \\
    \addlinespace
    Conceptual questions & Programming Language, Python Library, Computer Science, Programming Tools, Mathematics, Other Concept \\
    \addlinespace
    Contextual questions & Assignment Clarification, Code Explanation, Interpret Output, Other \\
    \addlinespace
    Verification & Verify Code, Verify Report, Verify Output, Other \\
    \addlinespace
    Provide context & Assignment Information, Error Message, Code, Other \\
    \addlinespace
    Off topic & Chit-Chat, Greeting, Gratitude, Other \\
    \bottomrule
  \end{tabular}
  \caption{Student dialogue-act schema from StudyChat \citep{mcnichols2026studychat}. Broad categories are used in our analyses; specific acts are shown for reference.}
  \label{tab:da_taxonomy}
\end{table}

\subsection{Scaffolding Level by Current Student Dialogue Act}

Table~\ref{tab:da_level} reports the mean scaffolding level following each broad student dialogue-act category.
Among instructional requests, mean scaffolding ranges from 3.27 for contextual questions to 3.72 for writing requests.
Writing requests therefore receive the highest average assistance, while off-topic turns receive substantially less task-relevant assistance.
Despite variation across request types, the major instructional categories remain concentrated at relatively high scaffolding levels.

\begin{table}[t]
  \centering
  \small
  \begin{tabular}{lrrr}
    \toprule
    \textbf{Current student DA} & \textbf{Mean $\bar{L}$} & \textbf{SD} & \textbf{$n$} \\
    \midrule
    Writing request       & 3.72 & 0.73 & 3{,}714 \\
    Editing request       & 3.57 & 1.05 &    336 \\
    Provide context       & 3.53 & 0.79 & 2{,}683 \\
    Conceptual questions  & 3.38 & 0.56 & 4{,}568 \\
    Verification          & 3.34 & 0.78 &    553 \\
    Contextual questions  & 3.27 & 0.76 & 2{,}566 \\
    Off topic             & 0.87 & 1.32 &    188 \\
    \bottomrule
  \end{tabular}
  \caption{Mean scaffolding level by current student dialogue-act category.}
  \label{tab:da_level}
\end{table}

\subsection{Within-Dialogue-Act Transition Patterns}

The main analysis shows that scaffolding level is associated with the student's next dialogue act.
Table~\ref{tab:within_da_full} reports the same transition distributions separately within four common current student dialogue-act categories.
Because observations at L0--L2 are sparse within individual dialogue-act categories, these levels are grouped in this analysis.

\begin{table}[t]
  \centering
  \footnotesize
  \setlength{\tabcolsep}{3pt}
  \begin{tabular}{llrrrrrr}
    \toprule
    \textbf{DA} & \textbf{Lv} & \textbf{ConQ} & \textbf{WrR} & \textbf{PrC} & \textbf{CtxQ} & $n$ & $\chi^2$ ($p$) \\
    \midrule
    \multirow{3}{*}{conc\_q}
      & L3   & 60.7 & 12.6 &  8.2 & 14.5 & 2672 & \multirow{3}{*}{\shortstack{117.0\\($<$.001)}} \\
      & L4   & 53.2 & 15.1 & 12.6 & 13.9 & 1850 & \\
      & L0--2 & 56.5 &  6.5 &  8.7 & 17.4 &   46 & \\
    \midrule
    \multirow{3}{*}{wr\_req}
      & L3   & 20.3 & 47.2 &  7.3 & 14.5 &  600 & \multirow{3}{*}{\shortstack{177.1\\($<$.001)}} \\
      & L4   & 16.3 & 48.0 & 16.4 & 13.5 & 2986 & \\
      & L0--2 &  3.2 & 56.5 & 10.5 &  9.7 &  124 & \\
    \midrule
    \multirow{3}{*}{pr\_ctx}
      & L3   & 16.9 & 20.4 & 37.7 & 16.7 &  812 & \multirow{3}{*}{\shortstack{50.1\\($<$.01)}} \\
      & L4   & 16.5 & 22.1 & 38.4 & 16.1 & 1730 & \\
      & L0--2 & 13.5 & 14.9 & 53.9 &  8.5 &  141 & \\
    \midrule
    \multirow{3}{*}{ctx\_q}
      & L3   & 29.0 & 18.1 & 12.8 & 33.3 & 1478 & \multirow{3}{*}{\shortstack{104.6\\($<$.001)}} \\
      & L4   & 24.0 & 19.9 & 17.8 & 29.5 &  973 & \\
      & L0--2 & 11.3 & 16.5 & 13.9 & 46.1 &  115 & \\
    \bottomrule
  \end{tabular}
  \caption{Distribution of the next student dialogue act (\%) by scaffolding level, within current student dialogue-act category. Minor categories are omitted, so rows do not sum to 100\%.}
  \label{tab:within_da_full}
\end{table}

The association between scaffolding and subsequent dialogue behavior is statistically significant within all four current request categories.
The largest contrast appears among writing requests: provide-context turns occur after 16.4\% of L4 responses compared with 7.3\% of L3 responses.
These within-category comparisons show that the overall association is not explained solely by broad differences in the type of student request.

\subsection{Provide-Context Responses}

We additionally examine whether the student's next turn is labeled provide\_context, which includes turns in which students supply assignment information, error messages, code, or other task-relevant context.

\begin{figure}[t]
  \centering
  \includegraphics[width=\columnwidth]{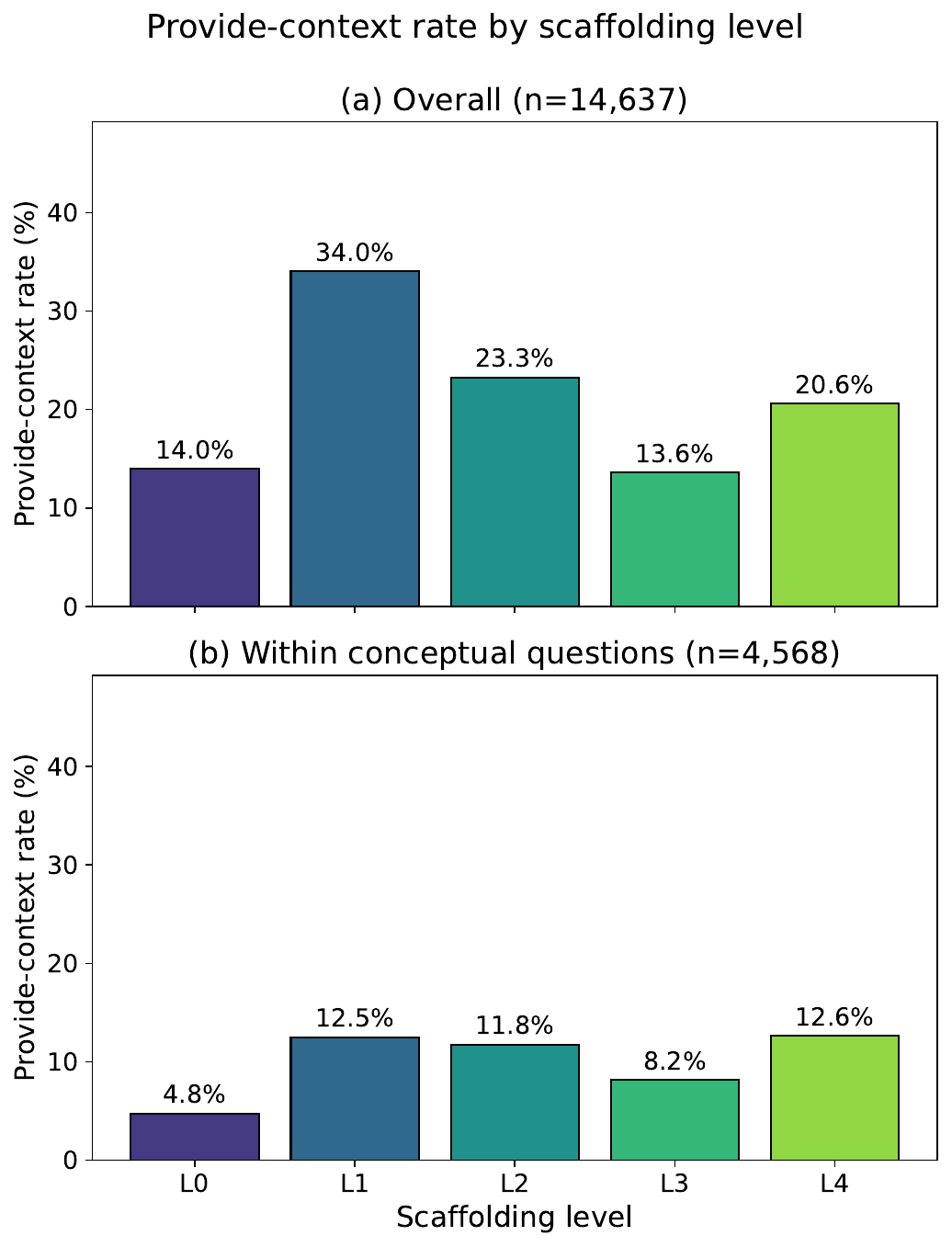}
  \caption{Provide-context rate by scaffolding level: (a) overall and (b) within conceptual questions.}
  \label{fig:provide_context}
\end{figure}

Across all interactions, the provide-context rate is 14.0\% after L0, 34.0\% after L1, 23.3\% after L2, 13.6\% after L3, and 20.6\% after L4.
Within conceptual questions ($n = 4{,}568$), the corresponding rates are 4.8\%, 12.5\%, 11.8\%, 8.2\%, and 12.6\% (Figure~\ref{fig:provide_context}).

The relationship is not monotonic.
In particular, within conceptual questions, provide-context turns occur at similar rates after Prompting and Solving responses.
Because the dialogue-act label indicates only that the student provides additional context, these data do not distinguish newly produced work from material adapted or copied from an earlier LLM response.
We therefore treat this measure descriptively rather than as a measure of independent student effort.

\subsection{Assessment Score Distributions}

Assignment scores exhibit strong ceiling effects, with an average assignment score of approximately .95 and substantial negative skew (Table~\ref{tab:score_dist}).
Exam scores show greater variation and were obtained without concurrent LLM access.
We therefore use the exams as the outcomes in the main RQ3 analysis.

\begin{table}[t]
  \centering
  \footnotesize
  \begin{tabular}{lrrrrr}
    \toprule
    \textbf{Assessment} & $n$ & \textbf{Mean} & \textbf{SD} & \textbf{Min} & \textbf{Skew} \\
    \midrule
    a1 & 181 & .986 & .032 & .867 & $-2.4$ \\
    a2 & 181 & .965 & .088 & .000 & $-8.1$ \\
    a3 & 181 & .894 & .097 & .145 & $-3.2$ \\
    a4 & 181 & .942 & .130 & .000 & $-5.5$ \\
    a5 & 181 & .962 & .082 & .000 & $-9.3$ \\
    a6 & 181 & .937 & .097 & .000 & $-5.6$ \\
    a7 & 181 & .952 & .114 & .000 & $-6.1$ \\
    \midrule
    e1 & 181 & .813 & .115 & .390 & $-1.0$ \\
    e2 & 181 & .928 & .126 & .000 & $-2.5$ \\
    e3 & 181 & .847 & .118 & .000 & $-2.5$ \\
    \bottomrule
  \end{tabular}
  \caption{Distribution of assignment and exam scores.}
  \label{tab:score_dist}
\end{table}

\section{Alternative Exam-Prediction Specifications}
\label{sec:alt_regression}

The main RQ3 specification uses mean scaffolding level $\bar{L}_{i,{<}a}$ as the scaffolding predictor.
We examine whether predictive information instead concentrates at either extreme of the scaffolding distribution by replacing mean scaffolding with two alternative predictors: the number of L4 (Solving) responses and the number of L0--L2 (low-assistance) responses received before each exam.
All other predictors remain unchanged.
Because the main analysis focuses on exams, we report the corresponding alternative specifications for e1--e3 (Table~\ref{tab:alt_regression}).

\begin{table}[t]
  \centering
  \small
  \begin{tabular}{llrrr}
    \toprule
    \textbf{Exam} & \textbf{Scaffolding predictor} & $\Delta R^2$ & $\hat{\delta}$ & $p_F$ \\
    \midrule
    e1 & L4 count      & +.001 & +.001 & .746 \\
    e1 & L0--2 count   & +.000 & +.001 & .855 \\
    e2 & L4 count      & +.001 & +.001 & .688 \\
    e2 & L0--2 count   & +.003 & $-.003$ & .400 \\
    e3 & L4 count      & +.000 & +.000 & .865 \\
    e3 & L0--2 count   & +.007 & +.003 & .227 \\
    \bottomrule
  \end{tabular}
  \caption{Alternative exam-prediction specifications using counts of Solving and low-assistance responses.}
  \label{tab:alt_regression}
\end{table}

Neither alternative scaffolding predictor reaches conventional statistical significance for any exam.
Both representations add little explained variance, consistent with the main analysis using mean scaffolding level.

\end{document}